\documentclass{article}

\PassOptionsToPackage{numbers, compress}{natbib}

\usepackage[preprint]{neurips_2026}

\usepackage{caption}
\usepackage[utf8]{inputenc} 
\usepackage[T1]{fontenc}    
\usepackage{hyperref}       
\usepackage{url}            
\usepackage{booktabs}       
\usepackage{amsfonts}       
\usepackage{nicefrac}       
\usepackage{microtype}      
\usepackage{xcolor}         
\usepackage{orcidlink}
\usepackage{pifont}   
\usepackage{xcolor}   
\usepackage{wrapfig}
\newcommand{\cmark}{\ding{51}}
\usepackage{graphicx}
\usepackage{booktabs}
\usepackage{multirow}
\usepackage{amsmath}
\usepackage{enumitem} 

\title{AdaptSplat: Adapting Vision Foundation Models for Feed-Forward 3D Gaussian Splatting}

\author{%
  Mingwei Xing\thanks{Equal contribution \quad $^\dagger$Corresponding author},\;
  Xinliang Wang$^*$,\;
  Yifeng Shi$^\dagger$ \\
  Ke Holdings Inc. \\
  \texttt{\{xingmingwei001, wangxinliang008, shiyifeng003\}@ke.com}
}

\begin{document}

\maketitle

\begin{abstract}
This work explores a simple yet powerful lightweight adapter design for feed-forward 3D Gaussian Splatting (3DGS). Existing methods typically apply complex, architecture-specific designs on top of the generic pipeline of image feature extraction $\rightarrow$ multi-view interaction $\rightarrow$ feature decoding. However, constrained by the scale bottleneck of 3D training data and the low-pass filtering effect of deep networks, these methods still fall short in cross-domain generalization and high-frequency geometric fidelity.
To address these problems, we propose AdaptSplat, which demonstrates that without complex component engineering, introducing a single adapter of only 1.5M parameters into the generic architecture is sufficient to achieve superior performance. Specifically, we design a lightweight Frequency-Preserving Adapter (FPA) that extracts direction-aware high-frequency structural priors from the shallow features of a powerful vision foundation model backbone, and seamlessly integrates them into the generic pipeline via high-frequency positional encodings and adaptive residual modulation. This effectively compensates for the high-frequency attenuation caused by over-smoothing in deep features, improving the fitting accuracy of Gaussian primitives on complex surfaces and sharp boundaries.
Extensive experiments demonstrate that AdaptSplat achieves state-of-the-art feed-forward reconstruction performance on multiple standard benchmarks, with stable generalization across domains.
Code available at: https://github.com/xmw666/AdaptSplat.
\end{abstract}

\section{Introduction}

\begin{wrapfigure}{r}{0.55\columnwidth}
  \centering
  \vspace{-19pt}
  \includegraphics[width=0.55\columnwidth]{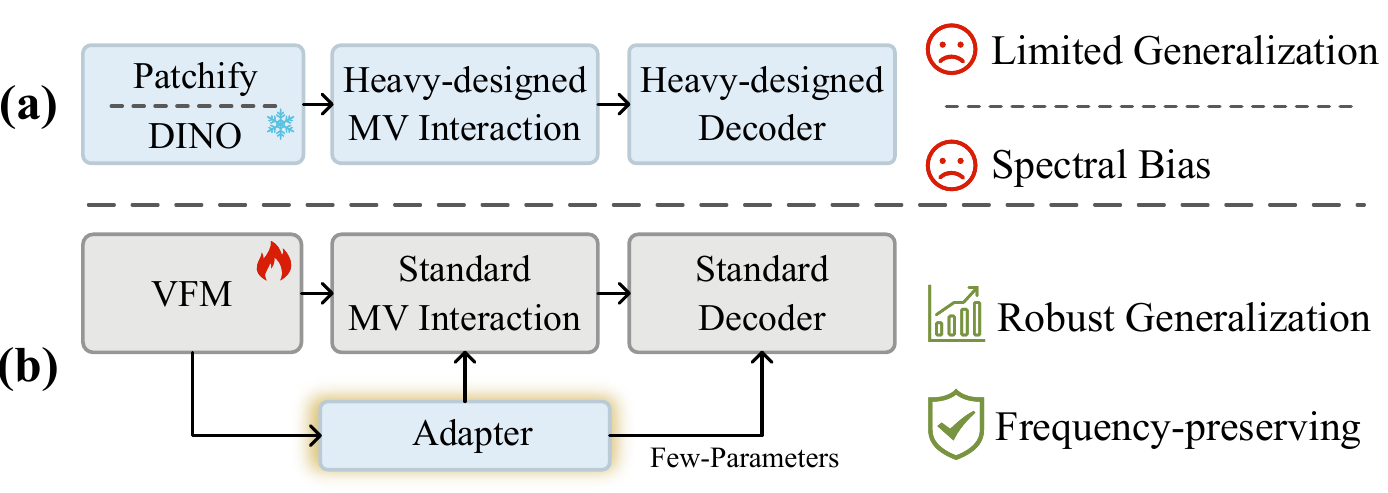}
  \caption{\textbf{Paradigm comparison.} Unlike \textbf{(a) existing methods} that struggle with weak generalization and spectral bias due to complex component designs, \textbf{(b) AdaptSplat} introduces a minimalist adaptation paradigm. It utilizes a single lightweight adapter to efficiently activate VFM priors, achieving superior generalization and high-fidelity reconstruction.}
  \label{fig:intro}
  \vspace{-17pt}
\end{wrapfigure}
Driven by the demand for scalable novel view synthesis, feed-forward 3D Gaussian Splatting (3DGS) has rapidly emerged as a dominant framework for generalizable scene reconstruction~\cite{globalsplat,long2026idesplat,jeong2026twoxplat}. To lift sparse 2D views into 3D representations, most existing methods have converged on a generic pipeline consisting of image feature extraction, multi-view interaction, and feature decoding. To improve reconstruction performance, prior work has invested substantial effort in designing complex task-specific architectures or heuristic training strategies for key modules within this pipeline, as shown in Figure~\ref{fig:intro} (a).

However, this prevailing paradigm of complex component engineering has several critical limitations. First, introducing intricate 3D-specific inductive biases into the architecture inevitably leads to complex design processes that rely heavily on manual priors. Second, constrained by the scale bottleneck of existing 3D training datasets, these over-engineered models often exhibit limited generalization to unseen scenes. To mitigate this, some methods incorporate 2D foundation models. However, existing approaches typically treat them as completely frozen feature extractors. This absolute freezing of parameters severs the active adaptation of general representations to multi-view 3D geometric constraints, limiting feature extraction capability and thus becoming a performance bottleneck for the entire pipeline. Third, deep neural networks inherently suffer from a low-pass filtering effect, causing high-frequency spatial details to be over-smoothed. Since inferring 3D geometry from 2D features is an ill-posed inverse problem, when the network is uncertain about local boundary directions due to edge smoothing, it tends to produce ``safe'' degenerate predictions---causing scaling coefficients to converge uniformly. This isotropic (spherical) degeneration prevents Gaussian primitives from accurately fitting complex object surfaces. Ultimately, there remains substantial room for improvement in the cross-domain generalization and high-frequency geometric fidelity of feed-forward 3DGS.

To this end, we propose AdaptSplat, as shown in Figure~\ref{fig:intro} (b), a simple yet powerful new paradigm for feed-forward 3DGS. We demonstrate that, without relying on complex task-specific pipeline redesign, introducing a tiny adapter of only 1.5M parameters into the generic architecture is sufficient to achieve superior performance. Specifically, we design a lightweight Frequency-Preserving Adapter (FPA). FPA is designed to fully unlock and adapt the robust multi-scale, multi-resolution representations of vision foundation models to 3D geometric constraints. It directly extracts direction-aware high-frequency structural priors from shallow features, and seamlessly injects them into the generic pipeline via high-frequency positional encodings and adaptive residual modulation. This effectively compensates for the high-frequency attenuation caused by over-smoothing in deep features, breaks the isotropic degeneration of Gaussian primitives, and significantly sharpens geometric boundaries.

Extensive experiments demonstrate that AdaptSplat achieves state-of-the-art feed-forward reconstruction performance on multiple standard benchmarks. Our contributions are as follows:
\begin{itemize}[leftmargin=*, itemsep=0pt, topsep=2pt, label=\textbullet]

\item \textbf{A minimalist adaptation paradigm for feed-forward 3DGS.} We abstract the generic pipeline of feed-forward 3DGS---image feature extraction, multi-view interaction, feature decoding---and demonstrate that without complex task-specific component engineering, introducing an ultra-lightweight adapter ($\sim$1.5M parameters) into this generic pipeline is sufficient to efficiently activate the strong generalization priors of vision foundation models, achieving comprehensive improvements in reconstruction performance.

\item \textbf{Frequency-Preserving Adapter (FPA) to break geometric degeneration.} To address the detail loss caused by the low-pass filtering effect of deep features, FPA directly extracts direction-aware high-frequency structural priors from shallow features. Through a dual injection mechanism of high-frequency positional encoding and adaptive residual modulation, it effectively compensates for high-frequency attenuation in features and significantly improves the fitting accuracy of Gaussian primitives on complex surfaces and boundaries.

\item \textbf{State-of-the-art performance and a minimalist new baseline.} AdaptSplat achieves state-of-the-art feed-forward reconstruction accuracy and superior cross-domain generalization on multiple benchmarks. We hope this approach can serve as a new baseline for feed-forward 3DGS, encouraging future research to shift focus from redundant pipeline design to efficient adapter engineering.

\end{itemize}

\section{Related Work}
\subsection{Feed-forward Gaussian Reconstruction}
\label{sec:21}
Building on 3D Gaussian Splatting (3DGS)~\cite{kerbl20233dgs,3dgs1,3dgs2,liu2025attentiongs,xu2025cruise,wang2026artifactworld,yogo}, recent research has progressively shifted toward feed-forward reconstruction from sparse views. Most methods have converged on a common pipeline: basic image patchification (e.g., MLPs/Convs) or vision foundation models (VFMs) for feature extraction, a multi-view Transformer~\cite{vggt} for cross-view interaction, and a decoder for Gaussian parameter regression.
However, the dominant trend is to apply complex task-specific modifications to each of these three components separately. On the feature extraction side, one line of work treats VFMs as strictly frozen feature extractors to transfer 2D semantic priors, but frozen representations cannot actively adapt to 3D geometric constraints, often requiring auxiliary strategies to compensate: DepthSplat~\cite{xu2025depthsplat} fuses monocular depth features; YoNoSplat~\cite{ye2026yonosplat} achieves pose-free reconstruction via a mix-forcing training strategy; VicaSplat~\cite{li2025vicasplat} jointly predicts 3D Gaussians and camera poses in a single forward pass; other methods~\cite{zhao2025erayzer,shi2025pmloss,jiang2025anysplat,noposplat,charatan2024pixelsplat,chen2024mvsplat,szymanowicz2024splatter} introduce generative priors or self-supervised learning. On the interaction and decoding side, another line of work aims to inject 3D inductive biases, replacing standard multi-view attention with epipolar transformers~\cite{transmvsnet,jia2025h3r} or cost volumes~\cite{xu2025depthsplat}, while progressively advancing the decoder from simple MLPs~\cite{chen2024longlrm} to DPT~\cite{mvp,ye2026yonosplat}, to better recover high-resolution spatial details.
Although these modifications improve geometric fidelity, they inevitably lead to constrained cross-domain generalization and spectral bias. In contrast, AdaptSplat directly adopts the generic pipeline described above, demonstrating that without dismantling or extensively customizing its components, inserting a single lightweight adapter of only 1.5M parameters is sufficient to address both problems simultaneously.

\subsection{Adapting Vision Foundation Models for 3D Vision}

Recent efforts explore parameter-efficient adaptation of vision foundation models to 3D tasks via lightweight adapters, preserving pretrained semantic priors and injecting geometric awareness without fully fine-tuning large backbones. MV-Adapter~\cite{mvadapter2024} and 3D-Adapter~\cite{3dadapter2024} introduce multi-view consistency modules and geometric feedback to improve cross-view alignment; Multi-View Foundation Models~\cite{multiviewfoundation2025} further integrate geometry-aware attention into pretrained encoders. For 3D understanding tasks, Image2Point~\cite{image2point2021} and CLIP2Point~\cite{clip2point2022} transfer 2D pretrained knowledge to 3D representations via lightweight adaptation modules; Adapt-As-You-Walk~\cite{adaptasyouwalk2025} demonstrates scalable adaptation of foundation models to 3D environments without retraining.
These approaches mainly focus on semantic transfer or cross-modal alignment, with little attention to generalization and reconstruction quality in feed-forward Gaussian reconstruction. AdaptSplat adapts the standard 3DGS reconstruction pipeline via FPA, simultaneously improving cross-domain generalization and high-frequency geometric fidelity.

\section{Method}

\subsection{Generic Feed-Forward 3DGS Pipeline}

Feed-forward 3DGS methods essentially follow a common pipeline: image feature extraction $\rightarrow$ multi-view interaction $\rightarrow$ feature decoding. A backbone network extracts visual features from input images; a multi-view Transformer handles cross-view interaction and geometric correspondence; a DPT decoder progressively restores spatial resolution; prediction heads decode features into Gaussian parameters $\mathbf{\mu}$, $\alpha$, $\mathbf{c}$, $\mathbf{s}$, $\mathbf{q}$, which are fed into a differentiable rasterizer for rendering. As noted in the introduction, existing methods face three core limitations on this pipeline: complex component engineering that relies heavily on manual priors, frozen VFMs that impede active adaptation and limit cross-domain generalization, and isotropic degeneration caused by the low-pass filtering effect of deep networks.

\subsection{AdaptSplat Overview}

To address the prevailing limitations of over-engineered pipelines, AdaptSplat adopts a minimalist adaptation paradigm. We begin by constructing a standard generic pipeline, intentionally omitting the complex 3D-specific inductive biases discussed in Section~\ref{sec:21}. Specifically, we use DINOv3-ConvNeXt~\cite{simeoni2025dinov3} as the feature extraction backbone, combined with a standard multi-view Transformer~\cite{vggt} for cross-view interaction, and a standard DPT decoder~\cite{dpt} for spatial feature regression.

The selection of DINOv3-ConvNeXt is driven by two practical merits. Architecturally, its hierarchical convolutions provide the multi-scale feature pyramids essential for extracting high-frequency priors. Operationally, its memory efficiency allows the VFM to be fully unfrozen for end-to-end training, directly adapting its semantic priors to 3D geometric constraints for robust generalization.

\begin{figure*}[t]
  \centering
  \includegraphics[width=0.85\linewidth]{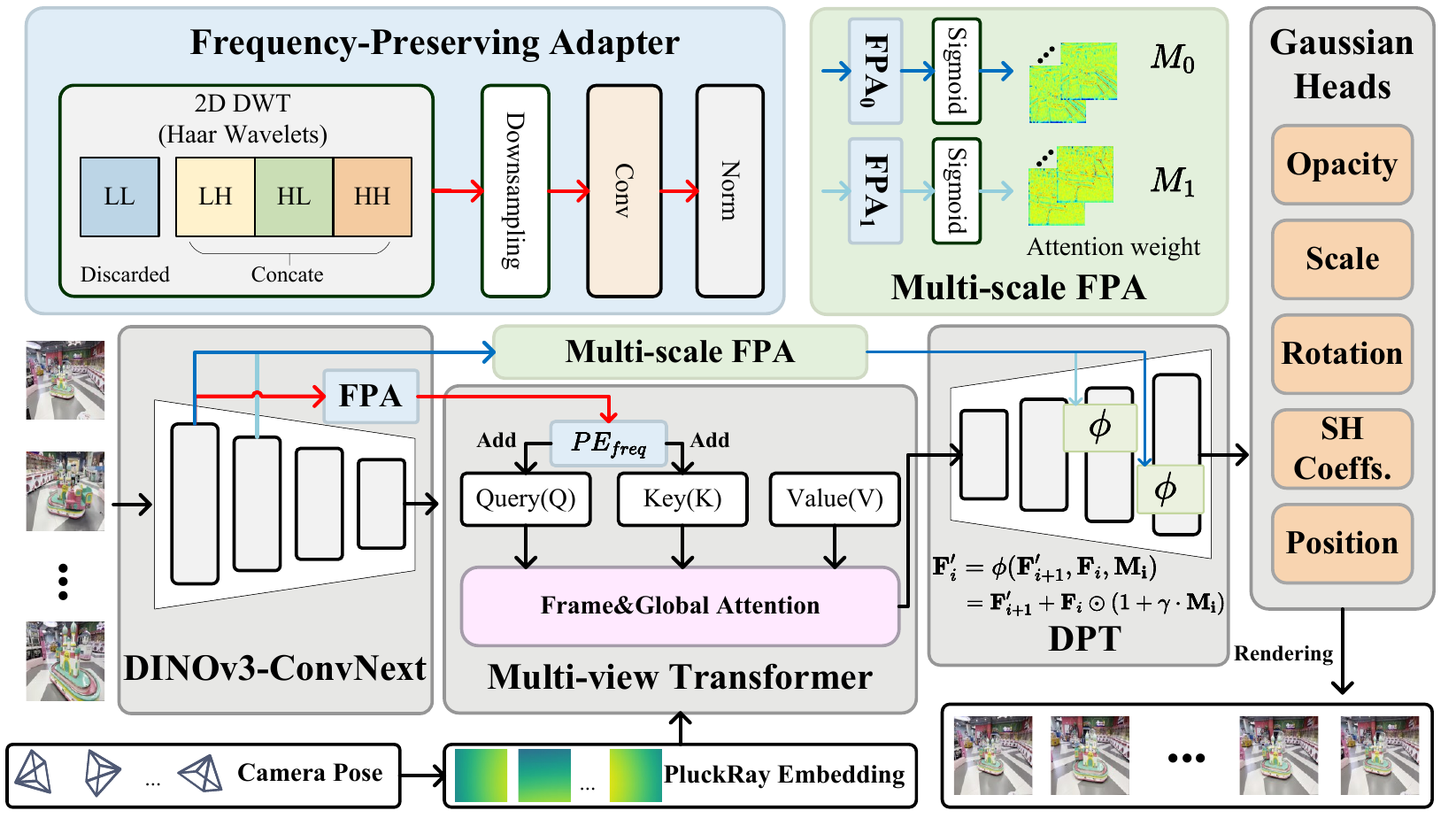}
  \caption{\textbf{Overview of AdaptSplat.} Based on the generic feature extraction-interaction-decoding pipeline, AdaptSplat introduces a lightweight Frequency-Preserving Adapter (FPA, ~1.5M parameters). FPA explicitly extracts high-frequency structural priors to combat the network's spectral bias. These priors are then injected into the Multi-view Transformer as frequency-guided positional encodings (PE) and into the DPT decoder via multi-scale adaptive residual modulation, significantly sharpening the 3D Gaussian primitives. }
  \label{fig:framework}
\end{figure*}

Instead of redesigning these core components, we introduce the lightweight Frequency-Preserving Adapter (FPA, ~1.5M parameters) as our primary customization (Figure~\ref{fig:framework}). FPA efficiently targets the deep networks' inherent low-pass filtering and isotropic degeneration. By focusing solely on this plug-and-play adapter, we prove that an unmodified generic pipeline, properly guided by VFM priors, is sufficient for state-of-the-art reconstruction. Section~\ref{sec:fpa} details the FPA.



\subsection{Frequency-Preserving Adapter (FPA)}
\label{sec:fpa}

In 3DGS, the geometry of a 3D Gaussian is determined by its covariance matrix $\Sigma = \mathbf{R}\mathbf{S}\mathbf{S}^\top\mathbf{R}^\top$. During differentiable rendering, its projected 2D covariance is $\Sigma' = \mathbf{J}\mathbf{W}\Sigma\mathbf{W}^\top\mathbf{J}^\top$, where $\mathbf{W}$ is the viewing transformation and $\mathbf{J}$ is the affine approximation Jacobian. Inferring 3D $\mathbf{S}$ and $\mathbf{R}$ from 2D features is a severely ill-posed inverse problem. Due to the low-pass filtering effect of deep networks, high-frequency boundaries are smoothed. When the network is uncertain about local boundary directions, it produces ``safe'' predictions: scaling coefficients converge to $s_x \approx s_y \approx s_z$, degenerating into isotropic Gaussian spheres.

FPA introduces 2D DWT to break this degeneration. DWT decomposes signals into $\mathbf{LL}$, $\mathbf{LH}$ (horizontal), $\mathbf{HL}$ (vertical), and $\mathbf{HH}$ (diagonal) subbands via orthogonal high-pass ($H$) and low-pass ($L$) filters. The $\mathbf{LH}$ and $\mathbf{HL}$ subbands capture high-frequency energy along orthogonal axes, providing a directional structure tensor for each region. This direction-aware guidance narrows the hypothesis space for $\mathbf{S}$ and $\mathbf{R}$, directing the network to perform anisotropic stretching along DWT-indicated boundary directions, breaking isotropic degeneration and enhancing high-frequency representation. A detailed quantitative analysis is provided in Section~\ref{sec:ablation}.

\subsection{High-Frequency Prior Injection}

The high-frequency priors $\mathbf{F}_{hf}$ extracted by FPA are injected into the generic pipeline via two mechanisms: guiding attention to perceive high-frequency structures in the multi-view Transformer, and resisting interpolation-induced high-frequency attenuation in the DPT decoder.

\subsubsection{High-frequency Guided Attention Positional Encoding.}
The self-attention mechanism in Transformers essentially performs a similarity-based Global Weighted Aggregation when updating features. This mechanism relies heavily on deep semantic correlations, yet lacks explicit perception of local high-frequency geometric boundaries, causing severe smoothing and blurring artifacts in rendering results.

To overcome this limitation, we abandon the conventional feature concatenation strategy and instead treat the high-frequency structural signals $\mathbf{F}_{hf}$ extracted by FPA as a form of positional encoding, explicitly injecting them into the Query ($\mathbf{Q}$) and Key ($\mathbf{K}$) spaces of self-attention. The modified attention computation is formulated as:
\begin{equation}
\text{Attention} = \text{Softmax}\left(\frac{(\mathbf{Q} + \mathbf{F}_{hf})(\mathbf{K} + \mathbf{F}_{hf})^\top}{\sqrt{d}}\right)\mathbf{V}
\end{equation}
The core of this design lies in achieving explicit decoupling of feature similarity computation from aggregation content. In the attention mechanism, the $\mathbf{Q}$-$\mathbf{K}$ space is responsible for computing inter-feature similarities to determine the allocation of attention weights, while the Value ($\mathbf{V}$) space carries the actual aggregated feature content. By injecting the high-frequency signals rich in directional priors exclusively into the $\mathbf{Q}$-$\mathbf{K}$ space, structural constraints are introduced into the similarity computation. This guidance encourages the network to preferentially aggregate within regions of similar structural features, thereby naturally maintaining sharpness in high-frequency regions. Meanwhile, this non-invasive injection strategy leaves the $\mathbf{V}$ space unmodified, perfectly preserving the clean semantic subspace of DINO's pre-trained features and fundamentally preventing interference from shallow high-frequency signals on its deep representations.

\subsubsection{High-frequency Adaptive Multi-scale Residual Modulation.}
Recovering feature resolution via bilinear interpolation is a common operation in the DPT decoding stage~\cite{mvp,ye2026yonosplat}. From a signal processing perspective, such spatial interpolation is essentially a low-pass filtering operation, causing secondary high-frequency attenuation of spatial details in feature maps. To address this upsampling degradation bottleneck, we leverage the high-frequency priors extracted by the FPA module and design a multi-scale spatially adaptive gating mechanism. Specifically, we pass the high-frequency features output by FPA through a Sigmoid activation function to generate dynamic spatial gating masks $\mathbf{M} \in (0,1)^{H \times W}$. The multi-scale features in the decoder then perform adaptive residual modulation:
\begin{equation}
\begin{aligned}
\mathbf{F}'_i &= \phi(\mathbf{F}'_{i+1}, \mathbf{F}_i, \mathbf{M}_i) \\
&= \mathbf{F}'_{i+1} + \mathbf{F}_i \odot (1 + \gamma \cdot \mathbf{M}_i)
\end{aligned}
\end{equation}
where $\mathbf{F}'_{i+1}$ is the upsampled deep semantic features, $\mathbf{F}_i$ is the shallow features, $\odot$ denotes element-wise multiplication, and $\gamma$ is a learnable scaling factor.

\subsection{Gaussian Regression and Optimization}

The decoder output is finally mapped to 3D Gaussian parameters through multiple lightweight prediction heads: opacity $\alpha$, scale factor $s$, rotation quaternion $q$, and spherical harmonic coefficients (SH). The position $\mu$ is obtained through backprojection of the predicted depth map $D$ combined with camera rays: $\mu = \mathbf{o} + D \cdot \mathbf{d}$.

\noindent \textbf{Loss Functions.} The model is trained end-to-end, minimizing the composite loss function $\mathcal{L}_{total} = \lambda_{rec} \mathcal{L}_{rec} + \lambda_{ffl} \mathcal{L}_{ffl} + \lambda_{reg} \mathcal{L}_{reg}$. For reconstruction fidelity, we combine pixel-level MSE loss with perceptual LPIPS loss to construct the reconstruction term $\mathcal{L}_{rec} = \mathcal{L}_{MSE} + \lambda \mathcal{L}_{LPIPS}$, constraining photometric accuracy and structural similarity. To counter the inherent spectral bias of neural networks and create synergy with the proposed FPA module, we introduce Focal Frequency Loss (FFL), denoted as $\mathcal{L}_{ffl}$. This loss dynamically increases the model's attention to high-frequency components by minimizing the frequency-domain distance between predictions and ground truth. Additionally, we apply opacity regularization, denoted as $\mathcal{L}_{reg}$.

\section{Experiments}
\subsection{Datasets and Implementation Details}

\noindent \textbf{Datasets.} We evaluate AdaptSplat on two primary datasets: DL3DV~\cite{ling2024dl3dv} and RealEstate10K (RE10K)~\cite{re10k}. For DL3DV, a large-scale dataset of diverse scenes averaging 250--350 frames each, we utilize its COLMAP-preprocessed~\cite{colmap1,colmap2} poses and standard data splits. For RE10K (67,477 train / 7,289 test videos), we follow the data splits of recent feed-forward models~\cite{wang2025volsplat} for fair comparisons.
To assess generalization capability beyond the training distribution, we conduct zero-shot inference on two challenging datasets without any fine-tuning: Tanks\&Temples~\cite{tandt}, featuring complex outdoor scenes with intricate geometric structures, and MipNeRF360~\cite{mipnerf}, containing unbounded real-world captures with challenging lighting variations.

\noindent \textbf{Implementation Details.} Following MVP~\cite{mvp}, we train on DL3DV using a progressive three-stage strategy. Stage 1 initializes at $480 \times 256$, predicting 12 target views from 32 inputs for 100k iterations (~4 days) to establish robust feature correspondences (LR: $10^{-5}$ for DINO, $10^{-4}$ elsewhere). Stage 2 increases resolution to $960 \times 540$ for detail refinement. Aided by memory optimization strategies, it predicts 6 target views from 32 inputs for 50k iterations (~4 days, uniform LR: $10^{-5}$). Stage 3 maintains $960 \times 540$ resolution but adopts variable input views (16--128) with dynamically adjusted target views to enhance generalizability (30k iterations, ~3 days, LR: $10^{-5}$). For RE10K, we resize and center-crop images to $256 \times 256$, predicting 8 target views from 6 inputs using the same settings as DL3DV Stage 1.
For camera poses, we use Plücker ray encoding as well as PRoPE~\cite{prope}.
All trainings are conducted on 32 NVIDIA H200 GPUs using AdamW, a cosine annealing schedule (3k warmup steps), and loss weights $\lambda_{rec} = 1.0$, $\lambda_{ffl} = 0.1$, and $\lambda_{reg} = 0.01$.

\begin{table}[t]
    \centering
    \caption{\textbf{Comparison on RE10K.} From 6 input views → 8 novel views, $256 \times 256$. }
    \label{tab:comparison_re10k}
    \setlength{\tabcolsep}{10pt}
    \begin{tabular}{lccc}
        \toprule
        Method & PSNR $\uparrow$ & SSIM $\uparrow$ & LPIPS $\downarrow$ \\
        \midrule
        PixelSplat~\cite{charatan2024pixelsplat} & 28.95 & 0.900 & 0.163 \\
        MVSplat~\cite{chen2024mvsplat}    & 29.13 & 0.924 & 0.091 \\
        TranSplat~\cite{kim2025transplat}  & 29.62 & 0.928 & 0.084 \\
        DepthSplat~\cite{xu2025depthsplat} & 29.68 & 0.925 & 0.087 \\
    VolSplat~\cite{wang2025volsplat} & 31.30 & 0.941 & 0.075 \\
        YoNoSplat~\cite{ye2026yonosplat} & 29.57 & 0.919 & 0.077 \\
        Long-LRM~\cite{chen2024longlrm} & 32.66 & 0.945 & 0.073 \\
        MVP~\cite{mvp} & 32.89 & 0.948 & 0.067 \\
        \textbf{Ours (tiny)} & 33.70 & 0.955 & 0.063 \\
        \textbf{Ours (base)} & \textbf{33.86} & \textbf{0.956} & \textbf{0.062} \\

        \bottomrule
    \end{tabular}
\end{table}
\begin{table*}[t]
\centering
\caption{\textbf{Quantitative results on DL3DV at high resolution ($960\times540$).} }
\label{tab:dl3dv_comparison1}
\setlength{\tabcolsep}{1pt}
\resizebox{\textwidth}{!}{%
\begin{tabular}{l ccc ccc ccc ccc}
\toprule
\multirow{2}{*}{Method} & \multicolumn{3}{c}{16 views} & \multicolumn{3}{c}{32 views} & \multicolumn{3}{c}{64 views} & \multicolumn{3}{c}{128 views} \\
\cmidrule(lr){2-4} \cmidrule(lr){5-7} \cmidrule(lr){8-10} \cmidrule(lr){11-13}
 & PSNR $\uparrow$ & SSIM $\uparrow$ & LPIPS $\downarrow$ & PSNR $\uparrow$ & SSIM $\uparrow$ & LPIPS $\downarrow$ & PSNR $\uparrow$ & SSIM $\uparrow$ & LPIPS $\downarrow$ & PSNR $\uparrow$ & SSIM $\uparrow$ & LPIPS $\downarrow$ \\
\midrule
3D-GS$_{30k}$~\cite{kerbl20233dgs} & 21.48 & 0.753 & 0.252 & 24.43 & 0.827 & 0.191 & 27.34 & 0.883 & 0.146 & 29.43 & 0.914 & 0.123 \\
\midrule
Long-LRM~\cite{chen2024longlrm} & 21.05 & 0.708 & 0.297 & 23.97 & 0.778 & 0.267 & 23.60 & 0.789 & 0.260 & 21.24 & 0.739 & 0.308 \\
iLRM~\cite{kang2025ilrm} & 21.92 & 0.748 & 0.316 & 24.30 & 0.803 & 0.256 & 24.44 & 0.819 & 0.240 & 22.98 & 0.807 & 0.249 \\
MVP~\cite{mvp} & 23.76 & 0.798 & 0.239 & 25.96 & 0.847 & 0.187 & 27.73 & 0.881 & 0.154 & 29.02 & 0.903 & 0.134 \\
Ours & \textbf{24.06} & \textbf{0.813} & \textbf{0.230} & \textbf{26.28} & \textbf{0.860} & \textbf{0.177} & \textbf{27.98} & \textbf{0.891} & \textbf{0.145} & \textbf{29.27} & \textbf{0.911} & \textbf{0.127} \\
\bottomrule
\end{tabular}%
}
\end{table*}

\subsection{Qualitative and Quantitative Comparison}

\noindent \textbf{Comparison on RE10K.} Following the evaluation protocol introduced in VolSplat~\cite{wang2025volsplat}, we evaluate the synthesis quality of eight novel views given six input views,as shown in Table~\ref{tab:comparison_re10k}. To ensure a fair comparison, the results for YoNoSplat~\cite{ye2026yonosplat} are generated via direct inference using official pre-trained weights, whereas Long-LRM~\cite{chen2024longlrm} and MVP~\cite{mvp} are retrained and reproduced using their official repositories under identical configurations. Experimental outcomes indicate that our approach achieves state-of-the-art performance across all evaluation metrics. Specifically, the base version of our model, Ours(base), reaches a PSNR of $33.86$, which notably exceeds recent leading baselines including VolSplat ($31.30$) and MVP ($32.89$). Furthermore, we develop a lightweight variant, Ours-tiny, utilizing the DINO-ConvNeXt (tiny) architecture. This version retains a PSNR of $33.70$ despite a reduced parameter count, illustrating a favorable trade-off between reconstruction accuracy and computational efficiency.

\begin{table}[t]
\centering
\caption{\textbf{Quantitative results on DL3DV at $280\times512$ resolution. }}
\label{tab:dl3dv_comparison2}
\resizebox{\linewidth}{!}{%
\begin{tabular}{l ccc ccc ccc}
\toprule
\multirow{2}{*}{Method} & \multicolumn{3}{c}{6v} & \multicolumn{3}{c}{12v} & \multicolumn{3}{c}{24v} \\
\cmidrule(lr){2-4} \cmidrule(lr){5-7} \cmidrule(lr){8-10} 
 & PSNR $\uparrow$ & SSIM $\uparrow$ & LPIPS $\downarrow$ & PSNR $\uparrow$ & SSIM $\uparrow$ & LPIPS $\downarrow$ & PSNR $\uparrow$ & SSIM $\uparrow$ & LPIPS $\downarrow$ \\
\midrule
MVSplat~\cite{chen2024mvsplat}    & 22.659 & 0.760 & 0.173 & 21.289 & 0.709 & 0.224 & 19.975 & 0.662 & 0.269 \\
DepthSplat~\cite{xu2025depthsplat} & 23.418 & 0.797 & \textbf{0.136} & 21.911 & 0.753 & 0.179 & 20.088 & 0.690 & 0.240 \\
YoNoSplat~\cite{ye2026yonosplat} & 24.717 & 0.817 & 0.139 & 23.285 & 0.773 & 0.177 & 22.664 & 0.758 & 0.192 \\
\textbf{Ours} & \textbf{25.795} & \textbf{0.847} & 0.149 & \textbf{26.741} & \textbf{0.864} & \textbf{0.135} & \textbf{26.901} & \textbf{0.871} & \textbf{0.132} \\
\bottomrule
\end{tabular}%
}
\end{table}

\begin{figure*}[t]
  \centering
  \includegraphics[width=0.85\linewidth]{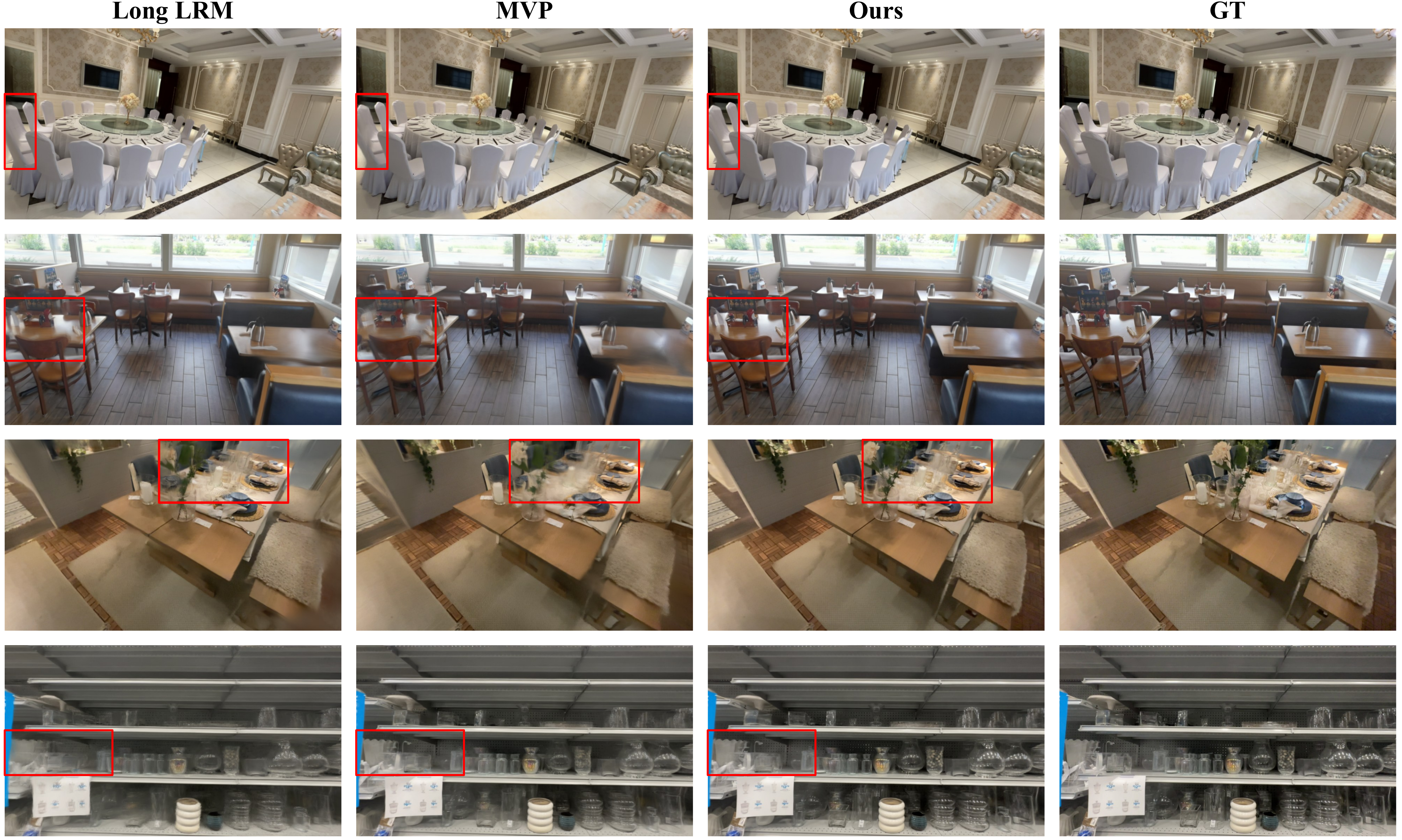}
  \caption{\textbf{Qualitative comparison on DL3DV.} AdaptSplat yields superior high-frequency fidelity and sharper geometric boundaries.}
  \label{fig:visual_dl3dv}
\end{figure*}

\begin{table*}[t]
\centering
\footnotesize
\setlength{\tabcolsep}{3pt}
\caption{\textbf{Zero-shot generalization on Tanks \& Temples and Mip-NeRF360.} Model trained on DL3DV, tested on unseen datasets without fine-tuning.}
\label{tab:comparison_dg}
\begin{tabular}{ll ccc ccc ccc}
\toprule
 & & \multicolumn{3}{c}{32 Views} & \multicolumn{3}{c}{64 Views} & \multicolumn{3}{c}{128 Views} \\
\cmidrule(lr){3-5} \cmidrule(lr){6-8} \cmidrule(lr){9-11}
Dataset & Method & PSNR $\uparrow$ & SSIM $\uparrow$ & LPIPS $\downarrow$ & PSNR $\uparrow$ & SSIM $\uparrow$ & LPIPS $\downarrow$ & PSNR $\uparrow$ & SSIM $\uparrow$ & LPIPS $\downarrow$ \\
\midrule
\multirow{4}{*}{Tanks \& Temples}
 & Long-LRM & 18.59 & 0.614 & 0.366 & 19.44 & 0.651 & 0.334 & 18.47 & 0.613 & 0.375 \\
 & iLRM     & 18.58 & 0.631 & 0.385 & 19.82 & 0.692 & 0.318 & 19.22 & 0.696 & 0.319 \\
 & MVP      & 19.54 & 0.708 & 0.277 & 21.24 & 0.761 & 0.221 & 22.36 & 0.804 & 0.184 \\
 & \textbf{Ours} & \textbf{19.80} & \textbf{0.731} & \textbf{0.265} & \textbf{21.80} & \textbf{0.789} & \textbf{0.207} & \textbf{22.81} & \textbf{0.826} & \textbf{0.171} \\
\midrule
\multirow{4}{*}{Mip-NeRF360}
 & Long-LRM & 21.08 & 0.484 & 0.445 & 21.30 & 0.499 & 0.431 & 19.82 & 0.484 & 0.457 \\
 & iLRM     & 21.09 & 0.495 & 0.466 & 21.60 & 0.522 & 0.444 & 21.32 & 0.551 & 0.424 \\
 & MVP      & 22.21 & 0.587 & 0.355 & 23.72 & 0.656 & 0.302 & 25.12 & 0.736 & 0.248 \\
 & \textbf{Ours} & \textbf{23.01} & \textbf{0.633} & \textbf{0.329} & \textbf{24.56} & \textbf{0.700} & \textbf{0.273} & \textbf{25.60} & \textbf{0.756} & \textbf{0.231} \\
\bottomrule
\end{tabular}
\end{table*}


\noindent \textbf{Comparison on DL3DV.} To evaluate model performance, we utilize the DL3DV dataset following the data partition criteria from MVP (Table~\ref{tab:dl3dv_comparison1}) and test reconstruction capabilities across a range of 16 to 128 input views. Quantitative evaluations show that AdaptSplat consistently outperforms state-of-the-art feed-forward baselines, such as MVP~\cite{mvp}, iLRM~\cite{kang2025ilrm}, and Long-LRM~\cite{chen2024longlrm}, achieving the highest scores across PSNR, SSIM, and LPIPS. This indicates that the proposed architecture effectively aggregates and leverages dense viewpoint information. Qualitative comparisons in Figure~\ref{fig:visual_dl3dv} highlight the visual superiority of our method. When handling intricate geometries like overlapping glassware or high-frequency textures on tabletops, Long-LRM and MVP suffer from noticeable blurring and structural degradation. Conversely, AdaptSplat produces sharp boundaries and clear local details by preserving and explicitly incorporating high-frequency signals, which yields results that closely match the ground truth.
Following the YoNoSplat~\cite{ye2026yonosplat} protocol (Table~\ref{tab:dl3dv_comparison2}), We further fine-tune the Stage 1 model at a resolution of $280 \times 518$, evaluating it with 6, 12, and 24 dynamic views. These view counts correspond to frame gaps of 50, 100, and 150. A higher number of views represents a larger spatial coverage and a longer camera trajectory. While the performance of baseline methods degrades as the scene scale and view count increase, AdaptSplat demonstrates a steady improvement. This trend confirms the capacity of our model to capture long-range features and maintain global geometric consistency in large-scale environments.

\noindent \textbf{Zero-shot Generalization.} To evaluate zero-shot generalization on unseen scenes, we directly apply the model trained exclusively on Stage 3 of the DL3DV dataset to the Tanks \& Temples~\cite{tandt} and Mip-NeRF360~\cite{mipnerf} datasets for inference (Table \ref{tab:comparison_dg}). We vary the number of input views from 32 to 128. Experimental results demonstrate that our model maintains superior reconstruction quality in unseen scenes without any task-specific fine-tuning. On the Mip-NeRF360 dataset in particular, the PSNR of our model increases steadily as the number of input views rises from 32 to 128, exceeding MVP by $0.8$, $0.84$, and $0.48$ dB, respectively. This performance demonstrates the architecture's robustness across varying visual densities. By integrating pre-trained DINO features—which effectively bridge domain gaps—our method achieves strong cross-dataset generalization, overcomes DL3DV training limits, and delivers state-of-the-art zero-shot rendering in the wild.

\subsection{Ablation Studies}
\label{sec:ablation}
\noindent \textbf{Component Ablation. }
To validate the effectiveness of each core component in the proposed methods, we perform ablation studies on a random 2k subset of the DL3DV dataset. All model variants are trained for 50k iterations at a $256 \times 480$ resolution to ensure a fair comparison. As shown in Table~\ref{tab:ablation}, the baseline model, which instantiates the generic pipeline with a frozen ConvNeXt encoder, yields limited performance with a PSNR of 21.12. Embedding DINOv3-ConvNeXt as a differentiable component within the optimization loop improves the PSNR to 21.47, a gain attributed to its robust semantic priors. Subsequently introducing FPA's high-frequency guided attention positional encoding further improves the PSNR to 21.75, demonstrating that the explicit injection of high-frequency directional priors effectively mitigates the spectral bias inherent in deep networks. Building upon these results, the FFL loss provides synergistic supervision in the frequency domain, further improving all evaluation metrics. Finally, incorporating multi-scale FPA in the decoder (M-FPA) enables adaptive residual modulation during upsampling, achieving the highest reconstruction fidelity with a PSNR of 22.10. These results underscore the necessity of their collaborative optimization.
\begin{table*}[t]
    \centering
    \begin{minipage}[t]{0.56\linewidth}
        \centering
        \caption{\textbf{Ablation study on DL3DV subset (2k scenes, 50k iterations)}}
        \vspace{5.5pt}
        \label{tab:ablation}
        \setlength{\tabcolsep}{2pt}
        \begin{tabular}{cccc|ccc}
            \toprule
            VFM & FPA & FFL Loss & M-FPA & PSNR $\uparrow$ & LPIPS $\downarrow$ & SSIM $\uparrow$ \\
            \midrule
            \multicolumn{4}{c|}{Baseline} & 21.12 & 0.3035 & 0.6813 \\
            \midrule
            \cmark &        &        &        & 21.47 & 0.2912 & 0.6992 \\
            \cmark & \cmark &        &        & 21.75 & 0.2683 & 0.7112 \\
            \cmark & \cmark & \cmark &        & 21.83 & 0.2675 & 0.7238 \\
            \cmark & \cmark & \cmark & \cmark & \textbf{22.10} & \textbf{0.2391} & \textbf{0.7321} \\
            \bottomrule
        \end{tabular}
    \end{minipage}
    \hfill
    \begin{minipage}[t]{0.40\linewidth}
        \centering
        \caption{\textbf{Comparison of high-frequency prior extraction strategies} on the DL3DV ablation subset.}
        \label{tab:hf_prior_extraction}
        \setlength{\tabcolsep}{4pt}
        \begin{tabular}{lccc}
            \toprule
            Method & PSNR $\uparrow$ & LPIPS $\downarrow$ & SSIM $\uparrow$ \\
            \midrule
            Base          & 21.47 & 0.2912 & 0.6992 \\
            Fourier       & 18.37 & 0.4767 & 0.5293 \\
            Conv          & 18.39 & 0.4805 & 0.5351 \\
            Sobel         & 21.45 & 0.2891 & 0.6091 \\
            \textbf{Ours} & \textbf{21.75} & \textbf{0.2683} & \textbf{0.7112} \\
            \bottomrule
        \end{tabular}
    \end{minipage}
\end{table*}

\begin{figure*}[tp]
    \centering
    \begin{minipage}[t]{0.7\linewidth}
        \centering
        \includegraphics[width=\linewidth]{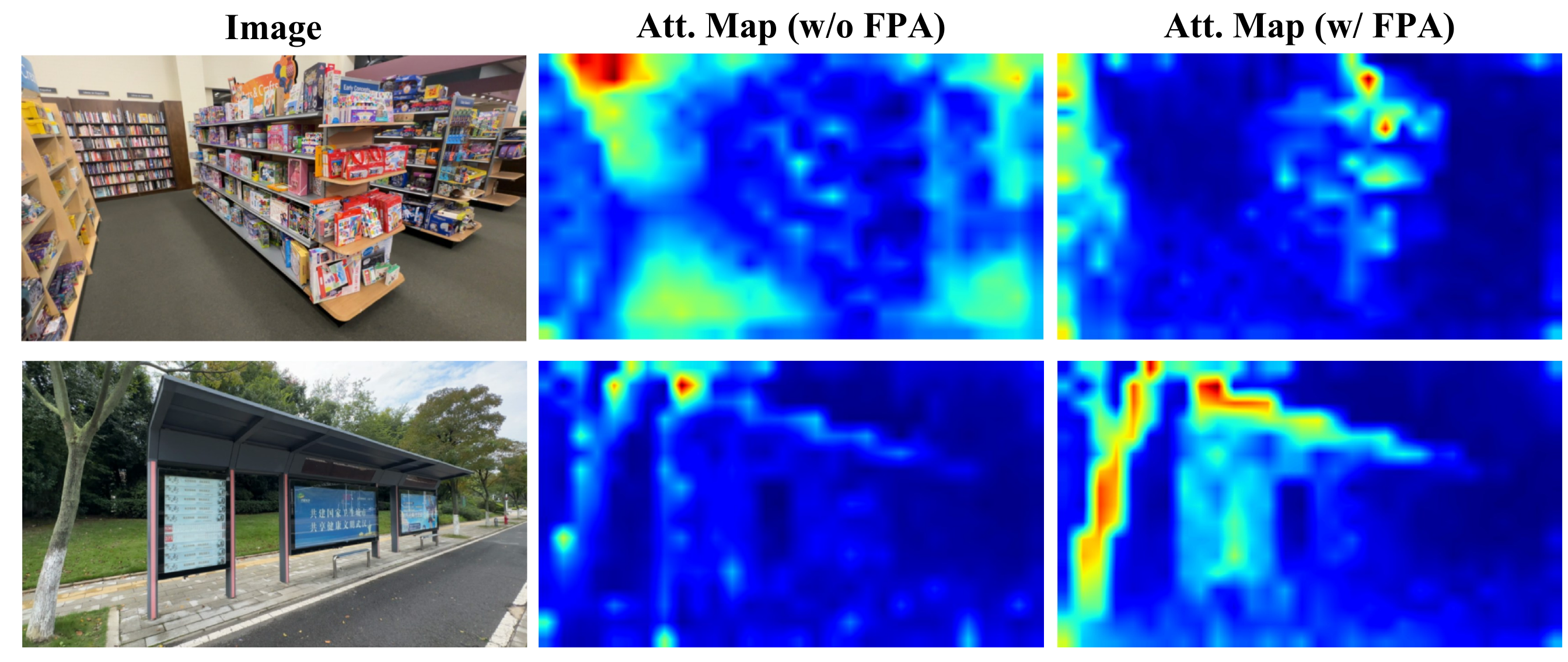}
        \captionof{figure}{\textbf{Attention maps Visualization.} Without FPA, attention is diffuse; with FPA, attention focuses on structural edges and feature boundaries are sharp.}
        \label{fig:att_pca}
    \end{minipage}
    \hfill
    \begin{minipage}[t]{0.28\linewidth}
        \centering
        \includegraphics[width=\linewidth]{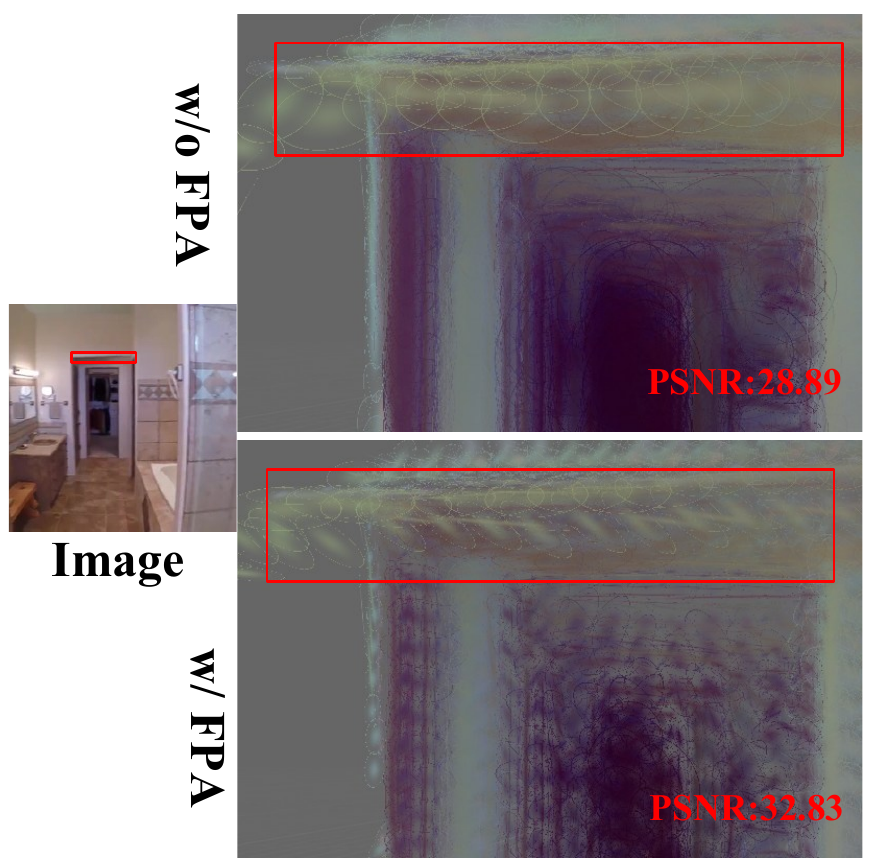}
        \captionof{figure}{\textbf{Gaussian distribution visualization at boundaries.} }
        \label{fig:edge_compare}
    \end{minipage}
\end{figure*}

\noindent \textbf{Frequency-Guided Attention Modulation.} We visualize attention maps within the Multi-view Transformer (Figure~\ref{fig:att_pca}). Due to the low-pass filtering effect of deep networks, the model without FPA produces diffuse attention: attention weights disperse into broad backgrounds and flat areas, while lacking concentration on the contours of key objects. This spatial ambiguity indicates that the baseline features lack local structural awareness. With FPA, the high-frequency priors from discrete wavelet transforms enable the attention mechanism to focus on structural edges and high-frequency texture regions, effectively suppressing spurious responses to low-frequency backgrounds.

\noindent \textbf{Anisotropic Gaussian Distribution Analysis.}
As analyzed in Section~\ref{sec:fpa}, inferring 3D covariance from 2D features is severely ill-posed. We visualize Gaussians at object boundaries to validate FPA's effectiveness in breaking this degeneration (Figure~\ref{fig:edge_compare}). Without FPA, Gaussians at edges appear as near-circular projections, indicating $s_x \approx s_y \approx s_z$ degeneration. With FPA, DWT-extracted directional priors guide the network to perform anisotropic stretching along boundary directions, producing elongated Gaussians that align with geometric structures.
To quantify this effect, we adopt the Fractional Anisotropy (FA) metric, which measures how much a tensor deviates from spherical:
\begin{equation} 
\bar{s} = \frac{s_x + s_y + s_z}{3}, \quad FA = \sqrt{\frac{3}{2}} \frac{\sqrt{(s_x - \bar{s})^2 + (s_y - \bar{s})^2 + (s_z - \bar{s})^2}}{\sqrt{s_x^2 + s_y^2 + s_z^2}},
\end{equation}

\begin{wraptable}{r}{0.33\linewidth}
    \centering
    \vspace{-18pt}
    \footnotesize
    \setlength{\tabcolsep}{5pt}
    \caption{\textbf{FA metrics on RE10K.}}
    \label{tab:fa}
    \begin{tabular}{lc}
        \toprule
        Method & FA $\uparrow$ \\
        \midrule
        w/o FPA & 0.8015 \\
        \textbf{w/ FPA (Ours)} & \textbf{0.8423} \\
        \bottomrule
    \end{tabular}
    \vspace{-15pt}
\end{wraptable}
where $s_x, s_y, s_z$ denote the three scale eigenvalues of each Gaussian covariance matrix. FA ranges from $0$ (perfectly isotropic sphere) to $1$ (fully anisotropic), with higher values indicating stronger directional stretching. As shown in Table~\ref{tab:fa} and Figure~\ref{fig:edge_compare}, FPA improves FA from $0.8015$ to $0.8423$, confirming that DWT-based directional priors effectively mitigate isotropic degeneration and produce anisotropic Gaussians aligned with geometric structures.

\noindent \textbf{Comparison of High-Frequency Prior Extraction Strategies. }
We compare alternative operators for constructing the high-frequency prior used by FPA (Table~\ref{tab:hf_prior_extraction}).
Besides our default wavelet decomposition, we consider frequency-domain filtering (Fourier), learnable high-pass convolutions (Conv), and Sobel edge responses (Sobel), all injected with the same adapter design.
Wavelet-based extraction achieves the best PSNR/SSIM and the lowest LPIPS, indicating that orthogonal scale--frequency separation is more reliable than hand-crafted or purely spectral alternatives.

\noindent \textbf{Comparison of Different Feature-level Adaptation Approaches.}
Given a fixed high-frequency prior, we further study how it is combined with the Multi-view Transformer expert (Table~\ref{tab:hf_expert_fusion}).
A simple channel-wise additive fusion (\textbf{Add}) slightly improves perceptual smoothness (lower LPIPS) but degrades global photometry and structure.
Injecting the prior as frequency-aware positional encodings on both queries and keys (\textbf{PE}) yields higher PSNR/SSIM by directly encoding frequency/positional correlations into the computation of attention weights.


\begin{figure*}[t]
    \centering
    \begin{minipage}[t]{0.35\linewidth}
        \centering
        \vspace{5pt}
        \captionof{table}{\textbf{Comparison of fusion schemes} between the high-frequency prior and multi-view transformer.}
        \label{tab:hf_expert_fusion}
        \setlength{\tabcolsep}{2pt}
        \begin{tabular}{lccc}
            \toprule
            Method & PSNR $\uparrow$ & LPIPS $\downarrow$ & SSIM $\uparrow$ \\
            \midrule
            Add & 21.16 & \textbf{0.2873} & 0.6868 \\
            PE (Ours) & \textbf{21.47} & 0.2912 & \textbf{0.6992} \\
            \bottomrule
        \end{tabular}
    \end{minipage}
    \hfill
    \begin{minipage}[t]{0.60\linewidth}
        \centering
        \captionof{table}{\textbf{Efficiency analysis on RE10K (6-view input).}}
        \label{tab:efficiency}
        \setlength{\tabcolsep}{1pt}
        \begin{tabular}{lcccc}
            \toprule
            Method & Param (M) & GPU Mem. (MB) & Time (s) & PSNR \\
            \midrule
            Long-LRM   & 141.92 & 3197  & 0.054 & 32.66 \\
            YoNoSplat  & 964.98 & 6437 & 0.677 & 29.57 \\
            MVP        & 241.23 & 4267  & 0.039 & 32.89 \\
            \midrule
            \textbf{Ours (tiny)} & 214.91 & 3127 & 0.042 & 33.70 \\
            \textbf{Ours (base)} & 419.98 & 4303 & 0.061 & \textbf{33.86} \\
            \bottomrule
        \end{tabular}
    \end{minipage}
\end{figure*}

\noindent \textbf{Efficiency Analysis. } As shown in Table~\ref{tab:efficiency}, Ours~(tiny) achieves $+0.81$ dB PSNR over MVP with fewer parameters and lower GPU memory. Ours~(base) outperforms YoNoSplat by $+4.29$ dB while using only $43\%$ of its parameters and running $11\times$ faster.

\section{Conclusion}
This paper presents AdaptSplat, a minimalist adaptation paradigm for feed-forward 3DGS. We abstract the generic feed-forward 3DGS pipeline and demonstrate that, without complex component engineering, introducing a lightweight adapter of only $\sim$1.5M parameters is sufficient to activate the generalization priors of vision foundation models and comprehensively improve reconstruction performance. The core module FPA extracts direction-aware high-frequency structural priors from shallow backbone features and injects them into the pipeline via a dual mechanism of high-frequency positional encoding and adaptive residual modulation, effectively compensating for high-frequency attenuation in deep features, breaking the isotropic degeneration of Gaussian primitives, and precisely fitting complex boundaries. Extensive experiments demonstrate that AdaptSplat achieves state-of-the-art reconstruction accuracy on multiple benchmarks with stable cross-domain generalization.

{
\small

\bibliographystyle{abbrvnat}

\bibliography{ref}
}
\newpage


\end{document}